\documentclass[conference]{IEEEtran}
\IEEEoverridecommandlockouts
\usepackage{pgfplots}
\usepackage{siunitx}
\pgfplotsset{compat=1.17}
\usepackage{cite}
\usepackage{amsmath,amssymb,amsfonts}
\usepackage{algorithmic}
\usepackage{graphicx}
\usepackage{textcomp}
\usepackage{xcolor}
\usepackage{diagbox}
\usepackage[utf8]{inputenc}
\usepackage[T1]{fontenc}
\usepackage[hidelinks]{hyperref}
\usepackage{booktabs}
\def\BibTeX{{\rm B\kern-.05em{\sc i\kern-.025em b}\kern-.08em
    T\kern-.1667em\lower.7ex\hbox{E}\kern-.125emX}}

\usepackage{multirow}

\usepackage[many]{tcolorbox}
\tcbset{myprompt/.style={subtitle style={colback={blue3!20}}}}

\definecolor{blue1}{rgb}{0,0.13,0.25}
\definecolor{blue2}{rgb}{0,0.22,0.42}
\definecolor{blue3}{rgb}{0,0.33,0.61}
\definecolor{blue4}{rgb}{0,0.4,0.75}
\definecolor{blue5}{rgb}{0,0.47,0.87}
\definecolor{blue6}{rgb}{0.11,0.56,0.95}

\newtcolorbox{taskbox}[2][]{%
    enhanced,breakable,
    colframe=blue3!40,
    colback=blue5!5,
    arc=1mm,
    outer arc=1mm,
    fontupper=\small,
    fontlower=\small,
    coltitle=blue1,
    fonttitle=\bfseries,    
    boxsep=1mm,
    left=0mm,
    right=0mm,
    top=0mm,
    bottom=0mm,
    before={\noindent},
    segmentation style={solid, blue3},
    title=#2,%
    #1
}

\title{Personalized Large Language Models}
    
\begin{document}


\author{\IEEEauthorblockN{
Stanisław Woźniak, Bartłomiej Koptyra, Arkadiusz Janz,\\ Przemysław Kazienko and Jan Kocoń}
\IEEEauthorblockA{\textit{Department of Artificial Intelligence, Wroclaw Tech, Poland}}
\IEEEauthorblockA{\texttt{\{stanislaw.wozniak, bartlomiej.koptyra, arkadiusz.janz,}} \IEEEauthorblockA{\texttt{przemyslaw.kazienko, jan.kocon\}@pwr.edu.pl}}
}

\maketitle

\begin{abstract}
Large language models (LLMs) have significantly advanced Natural Language Processing (NLP) tasks in recent years. However, their universal nature poses limitations in scenarios requiring personalized responses, such as recommendation systems and chatbots. This paper investigates methods to personalize LLMs, comparing fine-tuning and zero-shot reasoning approaches on subjective tasks. Results demonstrate that personalized fine-tuning improves model reasoning compared to non-personalized models. Experiments on datasets for emotion recognition and hate speech detection show consistent performance gains with personalized methods across different LLM architectures. These findings underscore the importance of personalization for enhancing LLM capabilities in subjective text perception tasks.
\end{abstract}

\begin{IEEEkeywords}
NLP, LLM, Personalization
\end{IEEEkeywords}

\section{Introduction}

In recent years, large language models (LLMs) have revolutionized Natural Language Processing (NLP) tasks in a variety of areas, demonstrating remarkable capabilities in text generation, sentiment analysis, machine translation, and more. These models are based on a transformer architecture \cite{vaswani2017attention} with a large number of parameters. Training such large models requires massive amounts of text data, enabling them to capture complex language patterns and generate consistent and contextually relevant text.

However, while LLMs have impressive generation capabilities, their universal nature is a limitation in scenarios where personalized responses are desired or required. Then, personalization becomes critical in applications such as recommendation systems, chatbots, and personalized content generation, where understanding and tailoring to individual subjective preferences and profiles is critical to user satisfaction and engagement.

Since the language models are zero-shot reasoners \cite{kojima2022large}, one can solve downstream tasks with prompt-based inference. In this way, we can obtain personalized answers by incorporating user-specific information into the instructions, i.e., in-context learning \cite{dong2022survey}. However, this approach does not update the model weights, so such personalization is impermanent, limited by the context length, and insufficient for specific downstream tasks. Another method we consider is to fine-tune the model in a personalized way. One of our goals is to investigate whether there is a difference between fine-tuning and zero-shot reasoning of LLMs on the subjective tasks.

Our contributions in this paper include: (1) a novel examination of fine-tuning versus zero-shot and few-shot reasoning in LLMs for personalizing subjective text perception, highlighting the superior performance of personalized fine-tuning; (2) a comprehensive evaluation across diverse datasets for emotion recognition and hate speech detection, demonstrating the significant advantages of personalization; (3) the proposal of new methods to enhance the LLM's responsiveness to individual user contexts, advancing subjective text analysis capabilities; (4) empirical validation of our approaches across various LLM architectures, underscoring their efficacy and adaptability; and (5) the release of research repository\footnote{ \url{https://github.com/Rikain/llm-finetuning}}, including code and datasets, to support reproducibility and further advancements in LLM personalization.

Our methodology is based on personalization through the use of basic user-specific context, which consists of user IDs. We utilized multiple fine-tuning approaches and few-shot in-context learning techniques to personalize LLMs for two distinct subjective tasks. Furthermore, we fine-tuned the models in both the classification and the text generation tasks. The results obtained in this work demonstrate that personalization methods based on simplified user-specific information, such as user IDs, have significant potential to enhance LLM performance by up to 165\%.

\section{Related Work}
AI has been increasingly applied to subjective tasks such as sentiment recognition, hate speech detection, and emotion recognition, leveraging techniques like deep learning and natural language processing. Models like transformers \cite{vaswani2017attention}, including BERT and GPT \cite{kenton2019bert,cambria2024senticnet}, have shown strong performance by capturing contextual information in text. However, these tasks remain challenging due to the ambiguity and variability in human language, often requiring large, well-labeled datasets to improve accuracy \cite{wierzba2021emotion,kocon2023deep}. Bias in training data and model fairness are also critical concerns, as they can affect the system's performance across different demographic groups. Despite these challenges, AI continues to advance in handling these nuanced tasks, showing promise in real-world applications.

Recent research highlights an interest in personalizing language models, emphasizing their significance across conversational interfaces, recommendation systems, and managing sensitive content\cite{porsdam2023autogen,lyu2023llm,abbasian2023conversational,chen2023palr,wu2023tidybot,zhu2024neurosymbolic}. Studies like \cite{kocon2021offensive,kanclerz2021controversy,kocon2021learning,milkowski2021personal,bielaniewicz2022deep,ngo2022studemo,kanclerz2022if,milkowski2022multitask,milkowski2023modeling,kanclerz2023towards,kanclerz2023pals,koptyra2023clarin,kocon2023multi,mieleszczenko2023capturing,ferdinan2023personalized,ferdinan2024fortifying} underscore the importance of tailoring NLP models to individual beliefs and preferences to enhance the handling of offensive content and controversial topics. Models that incorporate personal perspectives, as demonstrated in \cite{milkowski2021personal} and \cite{mireshghallah2022useridentifier},  offered superior predictions by acknowledging individual emotional responses.
Kazienko et al. \cite{kazienko2023human} extend this approach by developing deep learning models that account for individual differences, significantly outperforming traditional models in subjective tasks. Moreover, a study in \cite{kocon2023chatgpt} evaluates the performance of ChatGPT and GPT-4 in generating personalized responses, revealing that such customization improves predictive performance.

The bulk of this research has focused on adapting conventional neural network architectures, like LSTM and transformers (BERT, RoBERTa), for personalization in NLP, demonstrating the benefits of aligning models with user-specific characteristics, especially for sensitive or subjective content. However, recent exploration into large language models (LLMs) like ChatGPT and GPT-4, as noted in \cite{kocon2023chatgpt}, showcases their potential in few-shot scenarios without task-specific training, highlighting the advanced capabilities of LLMs to cater to individual user requests effectively.

Fine-tuning allows the models to adapt to specific downstream tasks, potentially leading to better performance. On the other hand, LLMs are sophisticated zero-shot reasoners \cite{kojima2022large}. One can use their abilities to solve downstream tasks with in-context-learning \cite{NEURIPS2020_1457c0d6} and extensive prompt-based inference \cite{yao2023tree}. Fine-tuning can be computationally expensive and time-consuming, especially for large language models. Fine-tuning a language model on task-specific data can improve its performance on the task, but it may come at the cost of reduced performance on other tasks. This is due to the risk of catastrophic forgetting \cite{french1999catastrophic}, where the model may forget some of the knowledge learned during pre-training and alignment processes \cite{li2024examining,kotha2023understanding,Zhai2023InvestigatingTC}. Techniques such as multi-task learning or balancing pre-training and task-specific data might be beneficial for retaining the performance of LLMs in multiple downstream tasks.

To the best of our knowledge, LLM fine-tuning for subjective tasks via user ID inputs, such as personalized emotion recognition or personalized hate speech detection, has not been extensively evaluated, and further research is needed in this area.

\section{Concept of Personalizing LLMs for Subjective Text Perception}

\begin{figure*}
    \centering
    \includegraphics[width=0.85\textwidth]{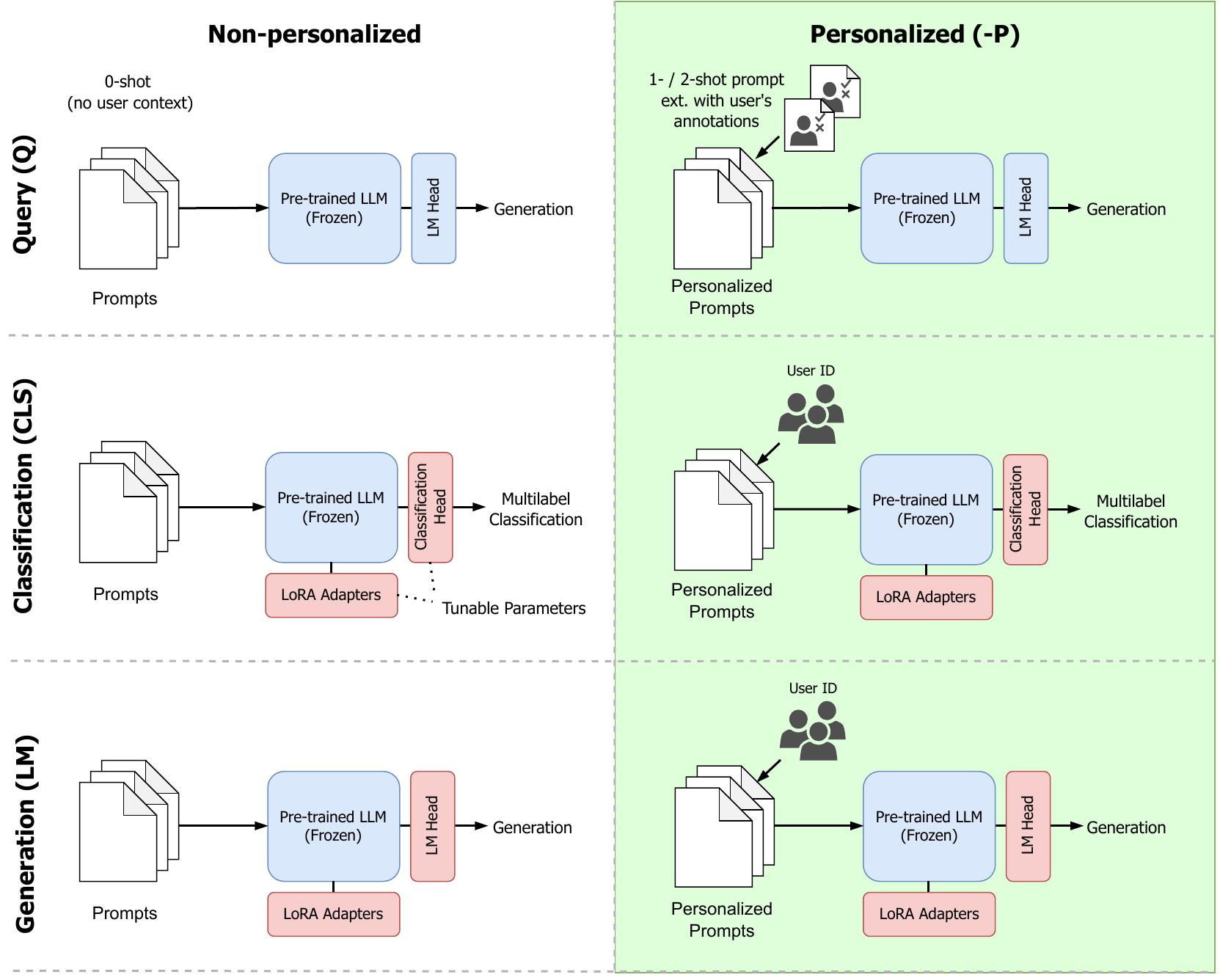}
    \caption{Non-personalized vs. personalized setups.}
    \label{fig:setups}
\end{figure*}

In human communication, interpretation and perception of texts depends not only on the textual content itself. For that reason, especially for subjective tasks like emotion recognition, hate speech, humor, or even sentiment analysis, LLMs should respect individual human preferences and beliefs, making their responses more personalized. Then, the models should be provided with information about the user either at the learning / fine-tuning stage (persistent) or at generation (impersistent). The concept of personalized LLM approaches (and non-personalized baselines) are presented in Fig.~\ref{fig:setups}.

\subsection{Problem Definition}
The primary challenge in personalizing LLMs for subjective text perception lies in the model's ability to incorporate individual user preferences, biases, and contexts into its processing mechanism. Given a user \(u\) and a text input \(T\), the goal is to generate a response $\hat{Y_u}$ that aligns with \(u\)'s subjective perception of \(T\). The prediction is defined as: 
\begin{equation*}
\hat{Y_u} = f(T, C_u)
\end{equation*}
where \(f\) is the function modeled by the LLM, and \(C_u\) represents the contextual user \(u\) information, which includes user preferences, historical interactions, and any other relevant user-specific data.

\subsection{Personalized Text Classification}
We propose a personalization approach that modifies the LLM's behavior based on \(C_u\). This can be achieved by adjusting the model's parameters \(\theta\) or by manipulating the input space to include personalized prompts. The objective function for personalization can be expressed as:
\begin{equation*}
\min_{\theta} \mathcal{L}(\theta; \hat{Y_u}, Y_u, T, C_u)
\end{equation*}
where \(\mathcal{L}\) is a loss function measuring the discrepancy between the generated response and a set of responses \(Y_u\) deemed appropriate by user \(u\). The personalization can be implemented through fine-tuning, where \(\theta\) is adjusted, or through in-context learning, where \(C_u\) is appended to \(T\) to guide the model's predictions.




\section{Non-personalized Baselines for Subjective Text Classification}

In evaluating the impact of personalization on LLMs, it is essential to establish non-personalized baselines. These baselines represent the model's performance without any adaptation to individual user contexts or preferences. We consider three primary non-personalized approaches: querying an original instruction-tuned model, classification with a new embedding head layer, and generative fine-tuning. Each method offers a different perspective on how LLMs handle subjective text classification in a non-personalized setting.

\subsection{Querying Instruction-tuned Language Models (Q)}
This approach involves utilizing a pre-trained LLM that has been instruction-tuned but not further adapted to any specific user data. Given a text input \(T\), the model generates a response \(\hat{Y}\) based on the instructions embedded during training:
\begin{equation*}
\hat{Y} = f_{\theta}(T)
\end{equation*}
where \(f_{\theta}\) represents the pre-trained LLM parameterized by \(\theta\), and \(T\) is the input text. This method evaluates the model's ability to follow instructions and generate appropriate classifications without any additional context or fine-tuning.

\subsection{Classification Head and Model Fine-tuning (CLS)}
In the classification approach, a new embedding head layer is introduced to the LLM for the specific task of text classification. The model is then fine-tuned. The objective function for fine-tuning can be defined as:
\begin{equation*}
\min_{\theta'} \mathcal{L}_{CLS}(\theta'; \hat{Y}, Y_u, T)
\end{equation*}
where \(\mathcal{L}_{CLS}\) is the classification loss function, \(\theta'\) are the parameters of the fine-tuned model including the new classification head, \(T\) is the input text, and \(Y_u\) represents user labels. This setup aims to adapt the LLM to perform the task better, still without considering any user-specific personalization.

\subsection{Generative Fine-tuning via Language Modeling (LM)}
Generative fine-tuning treats text classification as a text generation problem. The model is fine-tuned to generate a textual label as output, given a descriptive prompt and the input text: 
\begin{equation*}
\min_{\theta''} \mathcal{L}_{LM}(\theta''; \hat{L}, L_u, T)
\end{equation*}
where \(\mathcal{L}_{LM}\) is a loss function suitable for text generation tasks (e.g., cross-entropy loss summed over all positions in the sequence.), \(\theta''\) are the parameters of the fine-tuned generative model, \(T\) is the input text, and \(L_u\) is the textual label corresponding to \(T\). Unlike the classification approach, which directly predicts labels, this method generates labels as part of a continuous text output.

\section{Methods of LLM Personalization}\label{sec:methods}

Personalizing Large Language Models (LLMs) aims to tailor the model's responses to align with individual user preferences, history, and contextual nuances. This section outlines a formal description for various personalization techniques, including few-shot personalization, personalized classification, and personalized language modeling. These methods leverage user-specific data to enhance the model's relevance and accuracy in subjective text perception tasks.
\subsection{Few-shot Personalization (Q-NS)}
Few-shot personalization leverages a small number of examples to guide the model towards user-specific interpretations or responses. This technique involves modifying the input prompt to include \(N\) examples (\(E_1, E_2, \ldots, E_N\)) $\in E_u$ that reflect the user's texts and his perspective or preferences of these texts, which follows a typical In-Context-Learning setting. The input \(T\) with user context \(C_u\) are used to generate a response \(\hat{Y_u}\):
\begin{equation*}
C_u = (E_1, E_2, \dots, E_N)
\end{equation*}
\begin{equation*}
\hat{Y_u} = f_{\theta}(T, C_u)
\end{equation*}
where \(f_{\theta}\) is the pre-trained LLM parameterized by \(\theta\), \(T\) is the original input text, and examples \(E_i = (T_i', Y_i'),\; T_i' \neq T\) are the inputs that illustrate user-specific preferences $Y_i'$ for user-annotated texts $T_i'$. This method aims to prime the model with a context that mirrors the user's viewpoint, thereby personalizing its output.



\subsection{Personalized classification (CLS-P)}\label{sec:task-oriented}
Personalized classification adapts the LLM to specific users by incorporating user identifiers directly into the model's training process. This approach fine-tunes the model's parameters \(\theta\) to minimize the loss between the predicted labels and true labels, taking into account user-specific data. The objective function for personalized classification:
\begin{equation*}
\min_{\theta'} \mathcal{L}_{CLS-P}(\theta'; \hat{Y_u}, Y_u, T, C_u)
\end{equation*}
where \(\mathcal{L}_{CLS-P}\) is the personalized classification loss function, \(\theta'\) are the parameters of the personalized model, \(T\) is the input text, \(Y_u\) are the user labels, and \(C_u\) represents the contextual information (user ID) specific to the user \(u\). This method produces more accurate and personalized label predictions.



\subsection{Personalized Languge Modeling (LM-P)}\label{sec:generative}
In personalized language modeling, the goal is to fine-tune the LLM so that its generated text is tailored to the individual user's language use, preferences, or style. In our case, these are user labels in the text version. Like the classification approach, this method fine-tunes the model but focuses on generating personalized text outputs rather than predicting labels. The objective can be defined as:
\begin{equation*}
\min_{\theta''} \mathcal{L}_{LM-P}(\theta''; \hat{L_u}, L_u, T, C_u)
\end{equation*}
where \(\mathcal{L}_{LM-P}\) is the loss function for personalized language modeling, \(\theta''\) are the parameters of the fine-tuned generative model, \(T\) is the input text, \(L_u\) is the desired textual output for user $u$, and \(C_u\) contains the user-specific contextual information (user ID). This allows the model to generate relevant responses aligned with the user's preferences.

\section{Experiments}

We undertook a comprehensive set of experiments to rigorously evaluate our hypotheses, primarily focusing on multi-label classification tasks using several large language models. Our experimental design included all of the approaches described in section \ref{sec:methods}. However, in this section, we provide a detailed description of the experimental scenarios, explaining the models and datasets used to investigate the effectiveness of our methods. Additionally, we have used all the models and datasets described below for scientific purposes, which is in accordance with their licenses.

\subsection{Datasets}\label{sec:datasets}

In our experiments, we used two English-language datasets: GoEmotions \cite{demszky2020goemotions} and Unhealthy Conversations \cite{price2020six}. Both corpora encompass annotations contributed by numerous individual annotators, each reflecting their subjective perspectives and opinions in a multi-label classification task. Datasets were partitioned based on textual content, resulting in distinct training, validation, and test sets delineated by individual texts. Empty annotations were omitted from consideration. Each partition was refined to exclude outlier annotators, defined as those with annotation frequencies significantly deviating from the dataset's norm. Specifically, annotators contributing less than 5\% of annotations compared to the highest individual annotator were eliminated. Subsequently, the dataset was further refined to incorporate annotations from all annotators across each partition, ensuring comprehensive coverage of annotated data within each subset.
\subsubsection*{GoEmotions}
The GoEmotion dataset under the Apache-2.0 license comprises nearly 58k Reddit comments annotated by 82 distinct annotators, resulting in a total of over 211k individual annotations. Each annotator labeled the data using 28 unique emotional categories, including admiration, amusement, anger, annoyance, approval, caring, confusion, curiosity, desire, disappointment, disapproval, disgust, embarrassment, excitement, fear, gratitude, grief, joy, love, nervousness, optimism, pride, realization, relief, remorse, sadness, surprise, and neutral sentiment. Following the application of our data-cleaning procedure, the dataset was refined to include annotations from 72 annotators. This refinement yielded a training split containing over 146k samples, with validation and test splits each comprising over 18k samples.
\subsubsection*{Unhealthy Conversations}
The second dataset, termed "Unhealthy Conversation", encompasses approximately 230k annotations contributed by 588 annotators across more than 44k distinct online news comments. Each comment was categorized as either healthy or unhealthy, with additional annotations denoting seven attributes: antagonistic, hostile, dismissive, condescending, sarcastic, generalization, or unfair generalization. Following preprocessing procedures, the dataset was refined to comprise a training set consisting of roughly 168k samples, along with validation and test sets, each containing over 20k samples. In the refined iteration of the dataset, the number of annotators was reduced to 427. This dataset is published under the CC BY-NC-SA 4.0 license
\subsection{Models}\label{sec:models}

This study investigates the performance of small and moderate-sized language models developed by different research groups. The following models were selected for the experimental part.

\subsubsection*{Mistral}

Mistral 7B \cite{jiang2023mistral} is a language model under the Apache-2.0 license that outperforms previous models such as Llama across diverse benchmarks and approaches the coding performance of Code-Llama 7B \cite{roziere2023code}. With the use of grouped-query attention (GQA) and sliding window attention (SWA), Mistral 7B enhances performance and efficiency in comparison to even larger models such as Llama-2-13B \cite{touvron2023llama}.

\subsubsection*{Flan-T5}

Flan-T5 \cite{chung2022scaling} is an LLM based on the T5 text-to-text transformer model, which is a common encoder-decoder language modeling architecture. Flan-T5 is specifically designed for natural language understanding and generation tasks, showing the impact of parameter scaling and instruction-based fine-tuning. The model demonstrates that instruction fine-tuning scales effectively with both the number of tasks and the size of the model. In our experimental part, we used Flan-T5-XL model which has 3B parameters under Apache-2.0 license.\footnote{\url{https://huggingface.co/google/flan-t5-xl}}

\subsubsection*{Phi-2}

The Phi-2 model \cite{li2023textbooks} is a 2.7 billion-parameter small language model (SLM) published under MIT license that challenges the notion that bigger models are always better. It achieves reliable performance in reasoning and language understanding, outperforming models of greater size. This is attributed to different model scaling and the use of high-quality (textbook-quality) training data. Despite its smaller size, Phi-2 demonstrates great performance on specific benchmarks without the need for alignment through Reinforcement Learning from Human Feedback (RLHF) \cite{ouyang2022training}.\footnote{\url{https://huggingface.co/microsoft/phi-2}}

\subsubsection*{StableLM}

StableLM models include 3B and 7B parameter decoder-only language models, refined through fine-tuning on diverse chat and instruction-following datasets.\footnote{\url{https://huggingface.co/stabilityai/stablelm-tuned-alpha-3b}} Utilizing the NeoX transformer architecture \cite{black2022gpt}, these auto-regressive models are designed for chat-based applications.\footnote{\url{https://huggingface.co/EleutherAI/gpt-neox-20b}} In experiments, we utilised 3 billion-parameter model available at HuggingFace\footnote{\url{https://huggingface.co}}, where it was added under the license CC-BY-NC-SA 4.0.

\subsubsection*{ChatGPT}

ChatGPT is a group of models developed by OpenAI under OpenAI API license. Two of the models used in this research are GPT-3.5 and GPT-4. Currently, GPT-4 is one of the best models for multiple tasks, such as zero-shot reasoning. \footnote{\url{https://huggingface.co/spaces/lmsys/chatbot-arena-leaderboard}}

\subsection{Experimental setting}
Our experimental setup takes into account the perspective of both datasets (Sec.~\ref{sec:datasets}) and models (Sec.~\ref{sec:models}). We designed the settings to provide an understanding of the performance and effectiveness of personalized fine-tuning and in-context learning methods across different methods and subjective tasks.

Most of the LLMs selected for our study have a transformer architecture with a decoder-only configuration. Notably, these models required less computational resources for fine-tuning than models with an encoder-decoder architecture, such as Flan-T5. The experiments on language modeling methods with Flan-T5 model were omitted. In the language modeling setting (LM), we compared solely the decoder-only models. However, for the remaining methods, we compared its performance with Mistral 7B model, which has a decoder-only architecture more than twice the size of Flan-T5.

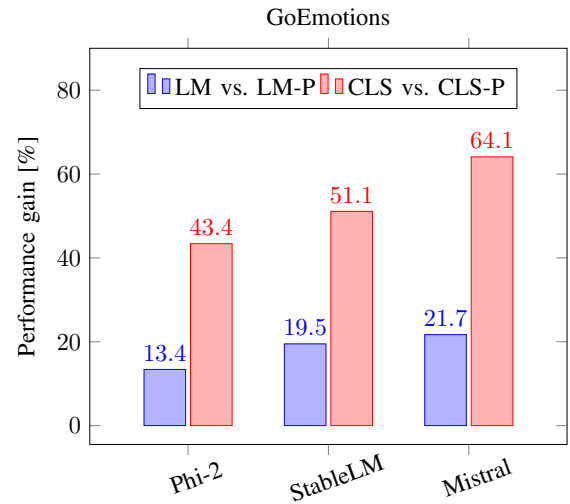
\begin{figure}
\centering
\resizebox{0.85\columnwidth}{!}{%
\begin{tikzpicture}
\begin{axis}[
    ybar,
    enlargelimits=0.35,
    enlarge y limits={value=0.05, lower},
    legend style={at={(0.5,0.95)},
      anchor=north,legend columns=-1},
    ylabel={Performance gain [\%]},
    symbolic x coords={Phi-2, StableLM, Mistral},
    xtick=data,
    nodes near coords,
    nodes near coords align={vertical},
    x tick label style={rotate=20,anchor=north,yshift=-1mm},
    title={GoEmotions},
    ymin=0,ymax=90,
    bar width=17pt,
]
\addplot coordinates {(Phi-2,13.4) (StableLM,19.5) (Mistral,21.7)};
\addplot coordinates {(Phi-2,43.4) (StableLM,51.1) (Mistral,64.1)}; 
\legend{LM vs. LM-P, CLS vs. CLS-P}
\end{axis}
\end{tikzpicture}
}
\caption{Performance gains of personalized vs. non-personalized methods on the GoEmotions dataset.\label{fig:pergoemo}}
\end{figure}

For fine-tuning and evaluation, our computational infrastructure consisted of four NVIDIA GeForce RTX 3090 GPUs, each with 24 GB of vRAM. Due to memory limitations, we employed modern fine-tuning techniques. We load the models using 4-bit NormalFloat (NF4) quantization and use qLoRA \cite{dettmers2023qlora} on all linear layers with the exception of the very last layer. In the case of CLS scenario the last layer is a newly initialised full layer, and in case of the LM task it is the LM head loaded in full precision. We do training in floating point 16-bit (fp16) for the StableLM and Mistral models and in BFloat16 (bf16) for the Flan-T5 and Phi-2 models. We made these implementations in Python using the PyTorch\footnote{\url{https://pytorch.org/}} library and HuggingFace libraries such as transformers\footnote{\url{https://huggingface.co/docs/hub/transformers}} and peft\footnote{\url{https://huggingface.co/docs/hub/peft}}. 

Given ChatGPT's remarkable performance in few-shot prompting, we wanted to examine its effectiveness in query methods (Q-0S vs. Q-1S and Q-2S) compared to the models from Section \ref{sec:models}. Our investigation included both versions of ChatGPT: GPT-3.5 and GPT-4.

\subsubsection*{Prompts}\label{sec:prompt}
Since Large Language Models (LLMs) operate as prompt-based reasoners, the input data for each experiment consisted of an instruction specifying the task to be performed by the model. Each prompt also included the text to be classified and a list of labels from which the model was expected to select the appropriate ones. The prompts we used are shown in the appendix~\ref{sec:appendix}.

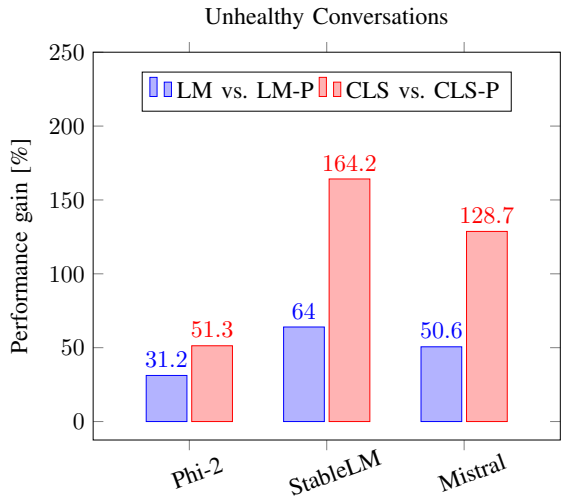
\begin{figure}
\centering
\resizebox{0.85\columnwidth}{!}{%
\begin{tikzpicture}
\begin{axis}[
    ybar,
    enlargelimits=0.35,
    enlarge y limits={value=0.05, lower},
    legend style={at={(0.5,0.95)},
      anchor=north,legend columns=-1},
    ylabel={Performance gain [\%]},
    symbolic x coords={Phi-2, StableLM, Mistral},
    xtick=data,
    nodes near coords,
    nodes near coords align={vertical},
    x tick label style={rotate=20,anchor=north,yshift=-1mm},
    title={Unhealthy Conversations},
    ymin=0,ymax=250,
    bar width=17pt,
]
\addplot coordinates {(Phi-2,31.2) (StableLM,64.0) (Mistral,50.6)};
\addplot coordinates {(Phi-2,51.3) (StableLM,164.2) (Mistral,128.7)};
\legend{LM vs. LM-P, CLS vs. CLS-P}
\end{axis}
\end{tikzpicture}
}
\caption{Performance gains of personalized vs. non-personalized methods on the Unhealthy Conversations.\label{fig:perunhealthy}}
\end{figure}


In fine-tuned personalized approaches (LM-P, CLS-P), we directly added the user ID to the prompt by inserting the line: \texttt{"\#\#\# User ID: <user ID>"}. Additionally, in the query (Q-0S, Q-1S, Q-2S) and language modeling (LM, LM-P) methods, we included a one-sentence request to the model within the prompt, specifying the expected format of the response: 

\emph{,,Please compose your response as a list of chosen labels, separated by commas.''}

In the Q-1S and Q-2S scenarios, we additionally included example texts and their correct responses within the prompt. Each example was separated from the instruction using a template on a new line: \texttt{"\#\#\# Example <N>: <example> \textbackslash n\#\#\# Example <N> Response: <example's response>"}. Where in the Q-1S scenario, the token \texttt{<N>} was left empty, while in the Q-2S scenario, \texttt{<N>} was replaced with the number 1 or 2, depending on the example.

\section{Results and discussion}

To compare the results across different approaches, we defined a \textit{gain} metric to quantify the percentage increase in quality of the personalized model relative to the baseline model:
\begin{equation*}
\textit{gain} = \left( \frac{\text{personalized} - \text{baseline}}{\text{baseline}} \right) \times 100\%
\end{equation*}
The results indicate that personalization of subjective task classification has a consistent impact on model performance, leading to significant performance gains (Fig.~\ref{fig:pergoemo} and Fig.~\ref{fig:perunhealthy}). This is empirically evident in personalized fine-tuning, i.e., CLS-P and LM-P settings vs CLS and LM. Moreover, fine-tuning within LM-P and CLS-P settings generally leads to better performance than zero-shot Q-0S and few-shot Q-1S, Q-2S settings. This suggests that while few-shot learning can adapt models to specific tasks without extensive retraining, fine-tuning remains a more effective strategy for maximizing model performance on specialized tasks.

The performance gains from personalization are more pronounced in the Unhealthy Conversations dataset than in the GoEmotions dataset -- see Table \ref{tab:query} and Table \ref{tab:fine-tuning}. This is in line with the research conducted in \cite{kocon2023chatgpt}, which shows that the optimal prompt-based personalization strategy (Q-1S and Q-2S settings) needs to be tailored to the specific characteristics and challenges of the task.

\begin{table}
\centering
\caption{\label{tab:fine-tuning}
The performance of large language models (LLM) on \emph{GoEmotion} and \emph{Unhealthy Conversations} datasets, computed for personalized settings (-P),  language modeling head (LM) vs. classification head (CLS). 
F1-macro scores are expressed in percentage points.
}
\resizebox{\columnwidth}{!}{
\begin{tabular}{lr|cccc}
\toprule
\multicolumn{2}{c|}{\diagbox{\textbf{Model}}{\textbf{Setting}}} & LM & LM-P & CLS & CLS-P \\
\midrule
& & \multicolumn{4}{c}{\textbf{GoEmotions}} \\[3pt]
\texttt{Phi-2} & 2.7B      & 28.99 & 32.87 & 30.03 & 43.07 \\ 
\texttt{StableLM} & 3B     & 26.55 & 31.72 & 27.42 & 41.44 \\ 
\texttt{Mistral} & 7B      & 28.36 & 34.52 & 26.77 & 43.94 \\
\midrule
& & \multicolumn{4}{c}{\textbf{Unhealthy Conversations}} \\[3pt]
\texttt{Phi-2} & 2.7B      & 34.97 & 45.89 & 31.91 & 48.26 \\ 
\texttt{StableLM} & 3B     & 29.61 & 48.54 & 16.92 & 44.68 \\ 
\texttt{Mistral} & 7B      & 34.29 & 51.65 & 23.10 & 52.83 \\
\bottomrule
\end{tabular}
}

\end{table}

The performance of few-shot settings varies across models and datasets, indicating that the effectiveness of few-shot learning might depend on the specific characteristics of models and the task. The GPT-family models from OpenAI, i.e., the GPT-3.5 and GPT-4, are the most consistent in few-shot settings -- the performance of Q-0S < Q-1S < Q-2S (see Table \ref{tab:query}) -- meaning they benefit more from an extended user context. On the other hand, the Mistral model does not fully utilize an extended user context in Q-1S and Q-2S few-shot settings despite undergoing an instruction fine-tuning procedure \cite{jiang2023mistral} as GPT-family models. Like Mistral, the instruction-based fine-tuning in Flan-T5 does not correspond well with personalized prompt-based approaches for subjective tasks.


The Phi-2 model was at a clear disadvantage with our prompts, as the model was not trained to follow instructions, not did it go through an alignment process known as Reinforcement Learning with Human Feedback (RLHF) or Direct Policy Optimization (DPO) \cite{rafailov2023direct}.

In our analysis, we observe a clear difference between Language Modeling (LM) and Classification (CLS) tasks, especially when considering their effectiveness in personalized settings across various datasets. Specifically, when dealing with the GoEmotions dataset, the personalized classification (CLS-P) method outperforms the personalized language modeling (LM-P) approach. This difference can be attributed to the GoEmotions dataset containing a wide range of labels, making it more challenging for language modeling techniques to capture subtle emotional nuances effectively. On the other hand, when evaluating the Unhealthy Conversations dataset, personalized language modeling (LM-P) shows notably better performance in one out of three experiments compared to personalized classification (CLS-P).

\begin{table}
\centering
\caption{\label{tab:query}
The performance of large language models (LLM) on \emph{GoEmotion} and \emph{Unhealthy Conversations} datasets, computed for baseline query with zero-shot setting (Q-0S) vs. personalized query with 1-shot (Q-1S), 2-shot (Q-2S) and fine-tuned language modeling head setting  (LM-P). 
F1-macro scores expressed in percentage points.
}
\resizebox{\columnwidth}{!}{
\begin{tabular}{lr|cccc}
\toprule
\multicolumn{2}{c|}{\diagbox{\textbf{Model}}{\textbf{Setting}}} & Q-0S & Q-1S & Q-2S & LM-P \\
\midrule
& & \multicolumn{4}{c}{\textbf{GoEmotions}} \\[3pt]
\texttt{Flan-T5} & 3B      & 17.97 & 17.79 & 17.87 & --- \\
\texttt{Mistral} & 7B      & 22.20 & 17.91 & 20.52 & 34.52 \\
\texttt{GPT 3.5} & - & 23.33 & 22.58 & 22.33 & --- \\
\texttt{GPT 4} & - & 26.74 & 26.82 & 26.64 & --- \\
\midrule
& & \multicolumn{4}{c}{\textbf{Unhealthy Conversations}} \\[3pt]
\texttt{Flan-T5} & 3B      & 14.11 & 12.81 & 12.68 &  --- \\
\texttt{Mistral} & 7B      & 18.90 & 26.12 & 23.21 & 51.65 \\
\texttt{GPT 3.5} & - & 21.51 & 23.11 & 25.49 & --- \\
\texttt{GPT 4} & - & 24.94 & 27.29 & 30.57 & --- \\
\bottomrule
\end{tabular}
}

\end{table}

 The performance differences between LM-P and CLS-P are less pronounced for Unhealthy Conversations than for GoEmotions. This is likely because Unhealthy Conversations has fewer labels, which suggests that label complexity has a significant impact on the effectiveness of personalized fine-tuning strategies. Tailoring fine-tuning approaches to the specific challenges presented by the task at hand is crucial. Dataset characteristics play a significant role in optimizing model performance.

Table \ref{tab:encoder-decoder} presents a comparison between Flan-T5, an encoder-decoder 3B parameter model, and Mistral, a decoder-only 7B model which is more than twice its size. Despite the potentially higher gains observed with the decoder-only model, the encoder-decoder architecture achieves better performance after fine-tuning (CLS vs CLS-P settings).

\begin{table}
\centering
\caption{\label{tab:encoder-decoder}
The performance of large language models (LLM) on \emph{GoEmotion} and \emph{Unhealthy Conversations} datasets, computed for baseline classification (CLS) vs. personalized classification (CLS-P) to show a comparison between our best model with decoder-only architecture and our model with encoder-decoder architecture. 
Metrics are expressed in percentage points.
}
\resizebox{\columnwidth}{!}{
\begin{tabular}{lr|cc|c}
\toprule
\multicolumn{2}{c|}{\multirow{2}{*}{\diagbox{\textbf{Model}}{\textbf{Setting}}}} & CLS & CLS-P & \multirow{2}{*}{Gain [\%]} \\
&& (F1-macro) & (F1-macro) &\\
\midrule
& & \multicolumn{3}{c}{\textbf{GoEmotions}} \\[3pt]
\texttt{Flan-T5} & 3B      & 32.64 & 45.68 & 39.95 \\
\texttt{Mistral} & 7B      & 26.77 & 43.94 & 64.14 \\

\midrule
& & \multicolumn{3}{c}{\textbf{Unhealthy Conversations}} \\[3pt]
\texttt{Flan-T5} & 3B      & 38.57 & 59.42 & 54.06 \\
\texttt{Mistral} & 7B      & 23.10 & 52.83 & 128.70\\
\bottomrule
\end{tabular}
}

\end{table}


\section{Conclusions and future work}

Our research underlines the crucial role of personalization in enhancing large language models (LLMs) for tasks involving subjective text perception. Through comprehensive experiments, we established that personalized fine-tuning significantly outperforms conventional zero-shot and few-shot learning methods, especially in the context of datasets with varying label complexities, such as GoEmotions and Unhealthy Conversations. The findings suggest that the success of personalization strategies is linked to the dataset's characteristics, underscoring the need for task-specific personalization approaches.

The study also reveals that LLMs' architecture and size critically influence the efficacy of personalization. Models like Mistral and the GPT family, which can follow detailed prompts and extended user contexts, show better improvements to models not specifically trained for instruction following or alignment through reinforcement learning.

Future research directions include examining the impact of personalization across a broader array of LLMs and subjective tasks and incorporating more contextual factors into personalization strategies. This could further enhance the precision and user-relevance of LLM outputs in personalized NLP applications.

\section*{Acknowledgment}
This work was financed by 
(1) the National Science Centre, Poland, project no. 2021/41/B/ST6/04471;  
(2) the statutory funds of the Department of Artificial Intelligence, Wroclaw University of Science and Technology;
(3) the Polish Ministry of Education and Science within the programme “International Projects Co-Funded”;
(4) CLARIN ERIC – European Research Infrastructure Consortium: Common Language Resources and Technology Infrastructure (period: 2024-2026) funded by the Polish Minister of Science under the programme: "Support for the participation of Polish scientific teams in international research infrastructure projects", agreement number 2024/WK/01;
(5) the European Union under the Horizon Europe, grant no. 101086321 (OMINO). However, the views and opinions expressed are those of the author(s) only and do not necessarily reflect those of the European Union or the European Research Executive Agency. Neither the European Union nor European Research Executive Agency can be held responsible for them.

\bibliographystyle{IEEEtran}
\bibliography{main}

\appendix

\section{Prompts}
\label{sec:appendix}

\begin{taskbox}[myprompt]{GoEmotions Prompts}
\tcbsubtitle{Prompt for Q-0S and LM scenarios}
Categorize the following text by selecting the most appropriate emotion from the provided list. Emotions can be subtle or overt, so analyze the text carefully to make an accurate classification. Please compose your response as a list of chosen emotions, separated by commas.\\
\\
\#\#\# Text:\\
<text>\\ 
\\
\#\#\# Emotions:\\
- <list of the all possible labels from hyphens>\\
\#\#\# Response:
\tcbsubtitle{Prompt for CLS scenario}
Categorize the following text by selecting the most appropriate emotion from the provided list. Emotions can be subtle or overt, so analyze the text carefully to make an accurate classification.\\
\\
\#\#\# Text:\\
<text>\\ 
\\
\#\#\# Emotions:\\
- <list of the all possible labels from hyphens>\\
\#\#\# Response:
\tcbsubtitle{Prompt for Q-1S scenario}
Knowing that for the given example was provided the response given below categorize the following text by selecting the most appropriate emotion from the provided list. Emotions can be subtle or overt, so analyze the text carefully to make an accurate classification. Please compose your response as a list of chosen emotions, separated by commas.\\
\\
\#\#\# Example:\\
<example text>\\
\\
\#\#\# Example Response:\\
<user's annotations for the example>\\
\\
\#\#\# Text:\\
<text>\\
\\
\#\#\# Emotions:\\
- <list of the all possible labels from hyphens>\\
\#\#\# Response:  
\tcbsubtitle{Prompt for Q-2S scenario}
Knowing that for the given examples were provided the responses given below categorize the following text by selecting the most appropriate emotion from the provided list. Emotions can be subtle or overt, so analyze the text carefully to make an accurate classification. Please compose your response as a list of chosen emotions, separated by commas.\\
\\
\#\#\# Example 1:\\
<first example text>\\
\\
\#\#\# Example 1 Response:\\
<user's annotations for the first example>\\
\\
\#\#\# Example 2:\\
<second example text>\\
\\
\#\#\# Example 2 Response:\\
<user's annotations for the second example>\\
\\
\#\#\# Text:\\
<text>\\
\\
\#\#\# Emotions:\\
- <list of the all possible labels from hyphens>\\
\#\#\# Response:  
\tcbsubtitle{Prompt for CLS-P scenario}
Categorize the following text for the specified user by selecting the most appropriate emotion from the provided list. Emotions can be subtle or overt, so analyze the text carefully to make an accurate classification.\\
\\
\#\#\# User ID:\\
<user ID>\\ 
\\
\#\#\# Text:\\
<text>\\ 
\\
\#\#\# Emotions:\\
- <list of the all possible labels from hyphens>\\
\#\#\# Response:
\tcbsubtitle{Prompt for LM-P scenario}
Categorize the following text for the specified user by selecting the most appropriate emotion from the provided list. Emotions can be subtle or overt, so analyze the text carefully to make an accurate classification. Please compose your response as a list of chosen emotions, separated by commas.\\
\\
\#\#\# User ID:\\
<user ID>\\ 
\\
\#\#\# Text:\\
<text>\\ 
\\
\#\#\# Emotions:\\
- <list of the all possible labels from hyphens>\\
\#\#\# Response:
\end{taskbox}

\begin{taskbox}[myprompt]{Unhealthy Conversation Prompts}
\vspace*{-0.12cm}
\tcbsubtitle{Prompt for Q-0S and LM scenarios}
Categorize the following text by selecting the most appropriate label from the provided list. Labels represent different types of communication styles or tones, where each category denotes a specific attitude or approach that someone might exhibit when communicating with others. Analyze text carefully to make an accurate categorization. Please compose your response as a list of chosen labels, separated by commas.\\
\\
\#\#\# Text:\\
<text>\\ 
\\
\#\#\# Labels:\\
- <list of the all possible labels from hyphens>\\
\#\#\# Response:
\tcbsubtitle{Prompt for CLS scenario}
Categorize the following text by selecting the most appropriate label from the provided list. Labels represent different types of communication styles or tones, where each category denotes a specific attitude or approach that someone might exhibit when communicating with others. Analyze text carefully to make an accurate categorization.\\
\\
\#\#\# Text:\\
<text>\\ 
\\
\#\#\# Labels:\\
- <list of the all possible labels from hyphens>\\
\#\#\# Response:
\tcbsubtitle{Prompt for Q-1S scenario}
Knowing that for the given example was provided the response given below categorize the following text by selecting the most appropriate label from the provided list. Labels represent different types of communication styles or tones, where each category denotes a specific attitude or approach that someone might exhibit when communicating with others. Analyze text carefully to make an accurate categorization. Please compose your response as a list of chosen emotions, separated by commas.\\
\\
\#\#\# Example:\\
<example text>\\
\\
\#\#\# Example Response:\\
<user's annotations for the example>\\
\\
\#\#\# Text:\\
<text>\\
\\
\#\#\# Labels:\\
- <list of the all possible labels from hyphens>\\
\#\#\# Response:  
\tcbsubtitle{Prompt for Q-2S scenario}
Knowing that for the given examples were provided the responses given below categorize the following text by selecting the most appropriate label from the provided list. Labels represent different types of communication styles or tones, where each category denotes a specific attitude or approach that someone might exhibit when communicating with others. Analyze text carefully to make an accurate categorization. Please compose your response as a list of chosen emotions, separated by commas.\\
\\
\#\#\# Example 1:\\
<first example text>\\
\\
\#\#\# Example 1 Response:\\
<user's annotations for the first example>\\
\\
\#\#\# Example 2:\\
<second example text>\\
\\
\#\#\# Example 2 Response:\\
<user's annotations for the second example>\\
\\
\#\#\# Text:\\
<text>\\
\\
\#\#\# Labels:\\
- <list of the all possible labels from hyphens>\\
\#\#\# Response:  
\tcbsubtitle{Prompt for CLS-P scenario}
Categorize the following text for the specified user by selecting the most appropriate label from the provided list. Labels represent different types of communication styles or tones, where each category denotes a specific attitude or approach that someone might exhibit when communicating with others. Analyze text carefully to make an accurate categorization.\\
\\
\#\#\# User ID: \\
<user ID>\\
\\
\#\#\# Text:\\
<text>\\ 
\\
\#\#\# Labels:\\
- <list of the all possible labels from hyphens>\\
\#\#\# Response:
\tcbsubtitle{Prompt for LM-P scenario}
Categorize the following text for the specified user by selecting the most appropriate label from the provided list. Labels represent different types of communication styles or tones, where each category denotes a specific attitude or approach that someone might exhibit when communicating with others. Analyze text carefully to make an accurate categorization. Please compose your response as a list of chosen labels, separated by commas.\\
\\
\#\#\# User ID: \\
<user ID>\\
\\
\#\#\# Text:\\
<text>\\ 
\\
\#\#\# Labels:\\
- <list of the all possible labels from hyphens>\\
\#\#\# Response:
\end{taskbox}

\end{document}